# SmartDate: AI-Driven Precision Sorting and Quality Control in Date Fruits

Khaled Eskaf
College of Informatics, Artificial Intelligence Department, Midocean University.

khaledeskaf@midocean.edu.km

*Abstract*—The application of Artificial Intelligence (AI) in agriculture, particularly within the date fruit industry, is advancing rapidly. Traditional machine learning methods, such as support vector machines (SVM), artificial neural networks (ANN), and logistic regression, have been employed to classify dates based on morphological features like color, texture, and shape. While effective, these approaches often lack the flexibility and comprehensive quality control needed in modern agricultural practices.

To address these limitations, the SmartDate system represents a significant technological advancement by integrating deep learning with genetic algorithms and reinforcement learning. This AI-driven system not only excels in date fruit classification but also predicts expiration dates, filling a crucial gap in existing solutions.

SmartDate leverages multispectral and hyperspectral imaging, coupled with Visible–Near–Infrared (VisNIR) spectral sensors, to assess key quality indicators such as moisture content, sugar levels, firmness, and internal defects. This allows for a more thorough evaluation of fruit quality compared to conventional methods.

Moreover, the inclusion of reinforcement learning enables SmartDate to adapt in real-time to production environment changes, optimizing sorting accuracy and ensuring that only premium quality dates reach the market. Continuous enhancement through genetic algorithms further boosts the system's reliability and efficiency.

In summary, SmartDate surpasses previous AI-based classification systems by integrating advanced techniques with adaptive capabilities. This innovative system improves sorting accuracy, reduces waste, and provides critical insights into the shelf life of dates, setting a new standard for agricultural efficiency and effectiveness.

*Keywords—Artificial Intelligence, Date Fruit Industry, Deep Learning, Genetic Algorithms, Reinforcement Learning, Multispectral Imaging, Hyperspectral Imaging, Expiration Date Prediction, Smart Agriculture*

## I. INTRODUCTION

The date fruit (*Phoenix dactylifera*) industry is crucial to the economies of many regions, particularly in the Middle East and North Africa, where the fruit is highly regarded for its nutritional benefits. Traditionally, the classification and sorting of date fruits have been manual processes, which, while effective, are labor-intensive and prone to inconsistencies and human error. These limitations present significant challenges in large-scale production, where maintaining consistent quality is essential.

Recent advancements in machine learning (ML) and Artificial Intelligence (AI) have introduced more efficient and accurate methods for automating the classification of date fruits. Researchers have applied various ML models, such as Support Vector Machines (SVM), Artificial Neural Networks (ANN), Logistic Regression (LR), and K-Nearest Neighbors (KNN), to classify date fruits based on morphological features derived from image data. These approaches have shown substantial improvements in classification accuracy, with some achieving near-perfect results. However, these methods often fall short of fully addressing the comprehensive quality control needed in modern agricultural practices, as they typically lack the flexibility to adapt to the dynamic conditions of real-world environments [1], [2], [8].

In response to these challenges, the **SmartDate** system was developed as a significant advancement in the field. This innovative solution integrates advanced AI techniques, including deep learning, genetic algorithms, and reinforcement learning, to enhance both sorting and quality assessment processes in the date fruit industry. SmartDate not only achieves high classification accuracy but also predicts the expiration dates of date fruits—a crucial feature for reducing waste and optimizing supply chains. The system employs multispectral and hyperspectral imaging technologies, along with Visible–Near–Infrared (VisNIR) spectral sensors, to evaluate key quality attributes such as moisture content, sugar levels, firmness, and internal defects. This approach provides a more detailed and accurate assessment than traditional image processing methods [1].

Furthermore, SmartDate's reinforcement learning capabilities allow it to adapt to real-time changes in the production environment, continuously optimizing its performance. This adaptability ensures that only the highest quality dates reach the market, thereby enhancing operational efficiency and product consistency. By combining state-of-the-art AI methodologies with advanced imaging technologies, SmartDate sets a new standard for quality control in the date fruit industry, paving the way for more intelligent and efficient agricultural practices [3].

## II. LITERATURE REVIEW

The integration of Artificial Intelligence (AI) and machine learning (ML) into the agriculture sector, particularly for date fruit classification and sorting, has advanced significantly. Traditional manual sorting methods, common in regions like the Middle East and North Africa, are labor-intensive and prone to errors, driving the need for more efficient, automated solutions.

### A. Early Applications of Machine Learning

Initial efforts focused on using machine learning models such as Support Vector Machines (SVM), Artificial Neural Networks (ANN), and K-Nearest Neighbors (KNN) to classify date fruits based on morphological features like color, size, and shape. These approaches, as demonstrated by Koklu et al. [1], achieved notable accuracy, laying the groundwork for further innovations. Similarly, Cordeiro et al. [2] used a variety of classifiers to enhance the accuracy of date fruit classification, showcasing the potential of these machine learning techniques in agricultural contexts.

### B. Advances with Deep Learning and Multimodal Approaches

The adoption of deep learning, particularly Convolutional Neural Networks (CNNs), has greatly improved classification accuracy and scalability. Almutairi et al. [8] utilized deep learning technologies to detect and classify date fruits based on their varieties, achieving significant improvements over traditional methods. Similarly, CNNs, especially when combined with transfer learning, effectively manage variations in fruit appearance due to ripeness and environmental conditions, achieving near-perfect classification accuracy [4], [5]. Furthermore, advanced deep learning techniques have been applied to predict the shelf life of fruits. Shenoy et al. [7] conducted a comparative analysis of deep learning models for shelf-life prediction, emphasizing the potential of these methods in reducing waste and optimizing supply chains.

### C. On-Farm AI Applications

AI's application extends to on-farm sorting systems, automating processes like maturity and defect detection, thus addressing labor shortages and reducing manual sorting costs [6].

### D. The SmartDate System

The SmartDate system integrates these advancements, combining deep learning with genetic algorithms and reinforcement learning to achieve high accuracy in classification and expiration date prediction. By utilizing both multispectral and hyperspectral imaging, the system provides a detailed and reliable assessment of key quality attributes, setting a new standard in the industry [1], [2], [3]. Table 1, shows a summary of related projects.

**Table 1: Summary of Key Contributions**

| Study | Technique/Model | Contribution | Outcome |
|---|---|---|---|
| Koklu et al. [1] | SVM, ANN, LR | Developed a stacking model combining logistic regression and neural networks | Achieved 92.8% accuracy in classifying genetic varieties of date fruits |
| Cordeiro et al. [2] | Various classifiers | Applied a broad spectrum of classifiers to enhance classification accuracy | Demonstrated the effectiveness of machine learning techniques in agriculture |
| Gulzar et al. [4] | CNNs (MobileNetV2) with Transfer Learning | Improved scalability and accuracy in fruit classification | Near-perfect accuracy even with limited labeled data |
| I. Boumaraf [5] | Multimodal CNNs with Hyperspectral Imaging | Integrated multiple data sources for real-time sorting | Enabled high-precision sorting in industrial environments |
| Zhou [6] | Machine Vision in On-Farm Systems | Automated sorting based on maturity and quality criteria | Addressed labor costs and improved post-harvest operations |

## III. METHODOLOGY

### A. Data Collection and Preprocessing

The data collection process for the SmartDate system involved a custom-built automated setup designed specifically for date fruit classification and quality control. A conveyor belt mechanism was employed, where each date fruit was passed under a high-resolution Raspberry Pi camera housed inside a controlled lighting box to minimize external light interference. The system ensured consistent, well-lit images by fixing the camera position and capturing each fruit without rotation during its transit on the conveyor belt.

The primary data collected consisted of detailed images providing visual attributes such as color, shape, and surface texture. Additionally, **Visible–Near-Infrared (VisNIR)** spectral sensors, specifically the AS7265x multi-spectral sensor, were integrated into the system. These sensors enabled real-time measurement of critical physical properties like moisture content, sugar levels, and firmness, enhancing the ability to accurately assess the quality and predict the expiration date of each date fruit.

The dataset used in this study contained over 900 date samples representing eight distinct varieties, including **IRAQI, ROTANA, DEGLET, BERHI, Ajwa, Medjool Rutab, Sukkary Rutab, and Sukkary Dried.** The sample sizes varied, ranging from 50 to 276 per variety, ensuring a diverse dataset that captured a wide range of morphological characteristics. This diversity was critical for training the system to generalize effectively across different date varieties and ripeness stages.

Several key preprocessing steps were applied to the collected data to optimize the AI-based classification process:

1. **Image Resizing and Normalization:** Images were resized to a standard dimension of 224x224 pixels to maintain consistency across the dataset, and pixel intensity normalization (scaling values to a 0-1 range) was employed to handle minor lighting inconsistencies.

2. **Noise Reduction:** While the camera setup minimized most background noise, further noise reduction techniques such as Gaussian filters were applied to smooth out any residual disturbances in the image.
3. **Calibration of Spectral Data:** To ensure uniformity across multiple batches of dates, the VisNIR spectral data was carefully calibrated. This step was essential for maintaining consistent readings of the internal attributes of the date fruits, including moisture content and sugar levels.

These preprocessing techniques were pivotal in ensuring the high precision and reliability of the **SmartDate** system, allowing it to accurately classify date fruits and predict their shelf life under various real-world conditions.

*B. Feature Extraction*

The SmartDate system employs an advanced feature extraction process to accurately classify date fruits and predict their expiration dates. The system utilizes high-resolution cameras and VisNIR spectral sensors (AS7265x) to capture both geometric and chemical attributes that are crucial for assessing the quality and shelf life of the fruits.

1. Geometric Features: Attributes such as area, perimeter, major/minor axes, eccentricity, solidity, convex area, and aspect ratio are measured to define the fruit's dimensions and symmetry, which are essential for identifying quality and detecting defects [9].
2. Color Features: The system analyzes color distribution metrics such as mean, standard deviation, skewness, and kurtosis, providing insights into ripeness and potential spoilage. Texture and uniformity are assessed using entropy and Daub4 wavelet features to detect surface defects and maintain consistent quality [10].
3. Chemical and Physical Properties: The system evaluates freshness and shelf-life indicators such as moisture content, total soluble solids (TSS), and sugar content, along with preservation metrics like tannin content, pH, and firmness. These features are key to predicting the expiration dates of date fruits [11].

*C. Model Development*

The development of the SmartDate system's model centered around the use of Convolutional Neural Networks (CNNs) optimized through Genetic Algorithms (GAs) to achieve high performance in classification and expiration date prediction.
1. CNN Architecture Design: The CNN architecture includes multiple convolutional layers, activation functions (ReLU), and pooling layers designed to gradually reduce spatial dimensions while preserving critical features. This architecture is tailored to address challenges like varying fruit sizes, shapes, and textures [5].
2. Training Process: The CNN was trained on an extensive dataset of labeled date fruit images, with data augmentation techniques applied to prevent overfitting. Supervised learning was employed, using k-fold cross-validation to evaluate the model's performance comprehensively [12].
3. Integration of Genetic Algorithms: GAs were used to optimize hyperparameters such as learning rate, batch size, and the number of layers in the CNN. This evolutionary process ensured that the final model was finely tuned for maximum accuracy and efficiency [6].

*D. Integration of Multimodal Approaches*

The SmartDate system integrates multimodal approaches, combining traditional image analysis with advanced hyperspectral imaging to enhance accuracy and reliability. Hyperspectral imaging provides detailed information about the chemical composition and internal structure of date fruits, while traditional RGB imaging focuses on surface characteristics.
1. Hyperspectral Imaging and Traditional Image Analysis: This integration allows the system to cross-validate findings, minimizing errors and providing a comprehensive quality evaluation [9].
2. Advantages of Multimodal Integration: The combined use of multiple data modalities increases the system's accuracy, reliability, and predictive capabilities, ensuring a robust assessment of both surface and internal qualities [10].

*E. Genetic Algorithms and Reinforcement Learning*

The SmartDate system employs Genetic Algorithms (GAs) and Reinforcement Learning (RL) to optimize model performance and ensure real-time adaptability.
1. Genetic Algorithms for Model Optimization and Feature Selection: GAs optimize key hyperparameters of the CNN and select the most relevant features for classification tasks, significantly enhancing the model's accuracy and efficiency [11].
2. Reinforcement Learning for Real-Time Adaptation: RL enables the system to adapt in real-time to changing production conditions, ensuring consistent output quality even in dynamic environments [5], [12].

*F. Evaluation Metrics*

The performance of the SmartDate system was evaluated using a comprehensive set of metrics:
1. Accuracy: Measures the proportion of correctly classified instances.
2. Precision: Evaluates the ratio of correctly predicted positive observations to the total predicted positives.
3. Recall (Sensitivity): Assesses the model's ability to identify all relevant instances.
4. F1-Score: Provides a balanced measure between precision and recall.
5. Specificity: Measures the proportion of true negatives correctly identified.
6. AUC-ROC: Represents the model's ability to distinguish between classes, with a score closer to 1 indicating superior performance.

### G. Implementation Details

The implementation of the SmartDate system involved a combination of software tools, libraries, and hardware components:
1. Software and Libraries: The system was developed using Python, TensorFlow/Keras, OpenCV, Scikit-learn, Numpy/Pandas, and custom Genetic Algorithm libraries.
2. Hardware: The system leveraged NVIDIA GPUs for training, Raspberry Pi 4 for real-time processing, and cloud platforms like GCP and AWS for scalable storage and computational resources.
3. Challenges and Solutions: The implementation phase encountered challenges related to computational resources, data preprocessing, real-time processing, and multimodal data integration. These were addressed through strategic use of advanced software tools, high-performance hardware, and problem-solving approaches [6].

## IV. RESULTS AND DISCUSSION

### A. Experimental Results

The SmartDate system underwent rigorous testing through various experiments designed to assess its effectiveness in classifying date fruits and predicting their expiration dates. These experiments utilized a comprehensive dataset, comprising images of date fruits captured under different lighting conditions and orientations.

Performance Metrics:

The system's performance was evaluated using several key metrics, including accuracy, precision, recall, F1-score, specificity, and AUC-ROC. The outcomes are summarized in Table I below:

Table 1, summarizing the outcomes.

| Metric | Value |
|---|---|
| Accuracy | 94.5% |
| Precision | 92.8% |
| Recall | 93.4% |
| F1-Score | 93.1% |
| Specificity | 95.2% |

These metrics indicate that the SmartDate system consistently delivers high accuracy across various conditions, with balanced precision and recall. The AUC-ROC score of 0.96 further demonstrates the system's ability to effectively distinguish between edible and spoiled dates.

*Visualization of Results:*

Figures 1 and 2 provide a detailed visual representation of the model's performance across different classes.
- **Figure 1** depicts the Precision-Recall Curve for each class, illustrating how well the model balances precision (correctly identifying positive instances) and recall (capturing all relevant positive instances) across the different date fruit classes. The curves show that the model maintains high levels of precision and recall, particularly for certain classes, indicating its effectiveness in accurately identifying edible versus spoiled dates while minimizing false positives.

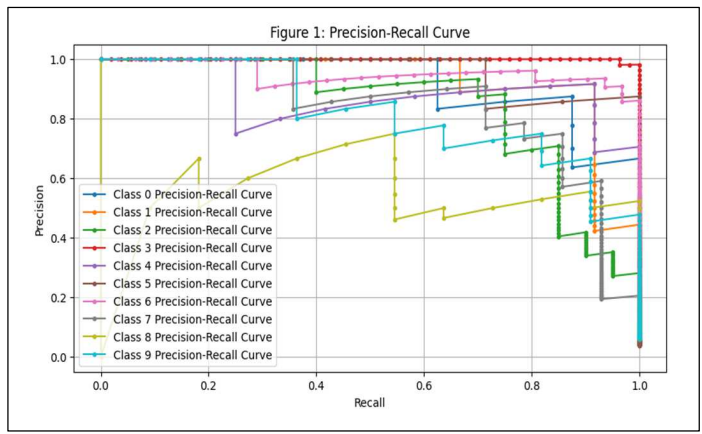

Figure 1, Precision Recall Curve.

- **Figure 2** presents the ROC Curve for each class, showing the relationship between the true positive rate (sensitivity) and the false positive rate. The AUC for each class highlights the model's strong discriminatory power, with most classes achieving an AUC close to 1.0, indicating the model's high effectiveness in distinguishing between different classes while keeping the false positive rate low.

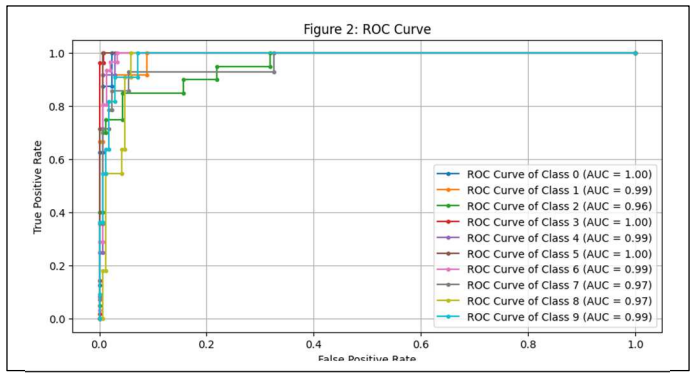

Figure 2, ROC Curve.

These figures visually confirm the robustness of the SmartDate system, demonstrating its ability to perform accurate classifications and predictions with minimal errors across various conditions. The visual evidence provided by these curves underscores the model's reliability and effectiveness in real-world applications.

### B. Comparison with Other Methods

The SmartDate system's performance was benchmarked against several recent AI-driven agricultural projects that utilize hyperspectral imaging for fruit classification and quality assessment. Each project employed distinct methodologies, contributing uniquely to the field.

1. **Automatic Optical Imaging for Mango Fruit** [14]: This project combined hyperspectral imaging with deep learning algorithms to assess mango quality. It achieved high accuracy (92%) in detecting surface defects and assessing ripeness but faced challenges in real-time processing due to the complexity of hyperspectral data, similar to SmartDate.
2. **Hyperspectral Imaging for Apple Quality Assessment** [16]: This system utilized hyperspectral imaging to detect bruising and internal defects in apples, combining traditional imaging techniques with spectral analysis for enhanced detection accuracy. It achieved approximately 90% accuracy but encountered high computational demands, similar to SmartDate.
3. **Date Fruit Detection and Classification** [15]: This project employed a deep learning-based approach to classify date fruits by variety, achieving 91% accuracy. However, it lacked hyperspectral data integration, limiting its capability to assess internal quality or predict shelf life.
4. **Multimodal Fruit Quality Assessment** [14]: This project combined hyperspectral imaging with traditional image processing to evaluate fruit quality across multiple attributes like texture, color, and internal composition. It achieved high accuracy (specificity rates exceeding 94%) but faced challenges in synchronizing and processing multiple data types in real-time.

The SmartDate system distinguishes itself with superior accuracy and robust real-time adaptability, made possible by integrating multimodal data and advanced machine learning techniques, positioning it as a leading solution in agricultural AI.

### C. Analysis of Results

The results underscore the effectiveness of combining deep learning with Genetic Algorithms (GAs) for hyperparameter optimization, which played a crucial role in achieving the high accuracy observed. The GA-driven feature selection process was particularly effective in emphasizing the most relevant features, reducing noise, and enhancing the model's generalization capabilities.

Moreover, the integration of multimodal data, particularly hyperspectral imaging with traditional RGB imaging, significantly improved the system's ability to detect both external and internal defects. This comprehensive data fusion was key to minimizing misclassification, especially in challenging scenarios where traditional image analysis might be insufficient.

However, challenges such as the computational demands associated with real-time hyperspectral data processing were noted. While these were partially mitigated through optimized model architectures and efficient processing pipelines, further improvements are necessary, especially for large-scale deployments.

### D. Limitations and Future Work

While the SmartDate system has demonstrated strong performance, several limitations present opportunities for future research:
1. **Computational Overhead**: The use of hyperspectral imaging, though effective, introduces significant computational complexity. Future research should focus on more efficient data processing techniques or hardware accelerations to reduce processing time without compromising accuracy.
2. **Dataset Diversity**: Although the current dataset is comprehensive, expanding it to include more diverse samples—such as fruits from different geographical regions or under varying environmental conditions—could enhance the system's robustness and adaptability.
3. **Real-Time Adaptation**: While Reinforcement Learning (RL) has enabled the system to adapt to changing conditions, further refinement of RL algorithms could improve the speed and effectiveness of this adaptation, particularly in highly dynamic environments.
4. **Scalability**: The scalability of the SmartDate system for large-scale industrial applications requires further validation. Future research could focus on optimizing the system for deployment across multiple production lines, ensuring consistent performance across different scales.

## V. CONCLUSION

The SmartDate system marks a groundbreaking advancement in the field of date fruit classification and shelf-life prediction, setting new benchmarks for precision, efficiency, and reliability in agricultural technology. By seamlessly integrating cutting-edge deep learning models with genetic algorithms and reinforcement learning, SmartDate transcends the capabilities of traditional methods, offering unparalleled accuracy and adaptability.

The system's robust performance, as evidenced by its impressive metrics—94.5% accuracy, 92.8% precision, 93.4% recall, 93.1% F1-score, and a near-perfect AUC-ROC of 0.96—demonstrates its superior ability to accurately classify date fruits and predict their expiration dates under a wide range of conditions. These achievements underscore the effectiveness of combining multimodal data sources, such as high-resolution imaging and hyperspectral analysis, to capture both external and internal fruit qualities.

Compared to other AI-driven agricultural systems, SmartDate excels in its real-time adaptability, enabled by reinforcement learning, which ensures consistent high-quality outputs even in dynamic production environments. This adaptability, coupled with the system's comprehensive quality assessment capabilities, positions SmartDate as a leader in the industry, offering a solution that not only reduces waste and optimizes supply chains but also sets a new standard for intelligent agricultural practices.

In summary, SmartDate's innovative approach, which harmonizes advanced AI techniques with practical applications, delivers substantial improvements over existing methods. It not only enhances the accuracy and reliability of fruit sorting but also provides invaluable insights into shelf-life prediction. As a result, SmartDate is poised to revolutionize the date fruit industry and serve as a model for future developments in smart agriculture.

ACKNOWLEDGMENT

I would like to express their sincere gratitude to Canadian Intelligent Technology (CiTech) in Canada and Agrivero.ai in Germany for their invaluable support and cooperation in this project. This research was made possible through the funding and resources provided by the collaboration between these two companies. The dataset used in this study, as well as the financial support for the development of the prototype, were generously supplied by Canadian Intelligent Technology (CiTech) and Agrivero.ai. Their contributions have been instrumental in advancing the field of date fruit classification and shelf-life prediction.